\documentclass[]{cas-dc}

\usepackage[numbers]{natbib}
\usepackage{graphicx}
\usepackage{amsmath,amssymb,amsfonts}
\usepackage{graphicx}
\usepackage{textcomp}
\usepackage{csquotes}
\usepackage{multirow}
\usepackage[normalem]{ulem}
\usepackage{longtable}
\usepackage{float}
\usepackage{adjustbox}
\usepackage{textcomp}

\begin{document}
\let\WriteBookmarks\relax
\def\floatpagepagefraction{1}
\def\textpagefraction{.001}
\shorttitle{Target specific mining of COVID-19 scholarly articles.}
\shortauthors{SK Sonbhadra et~al.}

\title [mode = title]{Target specific mining of COVID-19 scholarly articles using one-class approach}

\author[1]{Sanjay Kumar Sonbhadra}[orcid=0000-0002-7457-9655]
\cormark[1]
 \ead{rsi2017502@iiita.ac.in}
\credit{Conceptualization of this study, Methodology, Simulation}
\cortext[cor1]{Corresponding author}

\address[1]{IIIT Allahabad, Prayagraj, U.P. India-211015}

\author[1]{Sonali Agarwal}[orcid=0000-0001-9083-5033]
\ead{sonali@iiita.ac.in}
\author[1]{P. Nagabhushan}[orcid=0000-0002-4638-5482 ]
\ead{pnagabhuhsan@iiita.ac.in}

\credit{Writing - Original draft preparation}

\begin{abstract}
The novel coronavirus disease 2019 (COVID-19) began as an outbreak from epicentre Wuhan, People's Republic of China in late December 2019, and till June 27, 2020 it caused 9,904,906 infections and 496,866 deaths worldwide. The world health organization (WHO) already declared this disease a pandemic. Researchers from various domains are putting their efforts to curb the spread of coronavirus via means of medical treatment and data analytics. In recent years, several research articles have been published in the field of coronavirus caused diseases like severe acute respiratory syndrome (SARS), middle east respiratory syndrome (MERS) and COVID-19. In the presence of numerous research articles, extracting best-suited articles is time-consuming and manually impractical. The objective of this paper is to extract the activity and trends of coronavirus related research articles using machine learning approaches to help the research community for future exploration concerning COVID-19 prevention and treatment techniques. The COVID-19 open research dataset (CORD-19) is used for experiments, whereas several target-tasks along with explanations are defined for classification, based on domain knowledge. Clustering techniques are used to create the different clusters of available articles, and later the task assignment is performed using parallel one-class support vector machines (OCSVMs). These defined tasks describes the behaviour of clusters to accomplish target-class guided mining. Experiments with original and reduced features validate the performance of the approach. It is evident that the \textit{k}-means clustering algorithm, followed by parallel OCSVMs, outperforms other methods for both original and reduced feature space. 
\end{abstract}


\maketitle

\section{Introduction}

Coronaviruses are a large family of deadly viruses that may cause critical respiratory diseases to the human being. Severe acute respiratory syndrome (SARS) is the first known life-threatening epidemic, occurred in 2003, whereas the second outbreak reported in 2012 in Saudi Arabia with the middle east respiratory syndrome (MERS). The current outbreak is reported in Wuhan, China during late December 2019 \cite{stoecklin2020first}. On January 30, 2020, the world health organization (WHO) declared it a public health emergency of international concern (PHEIC) as it had spread to more than 18 countries \cite{SA1} and on Feb 11, 2020, WHO named this \enquote{COVID-19}. On March 11, 2020, as the number of COVID-19 cases reaches to 118,000 in 114 countries and over 4,000 deaths, WHO declared this a pandemic. 

Several research articles have been published on coronavirus caused diseases after 2003 till date. These articles belong to diverse domains like medicine and healthcare, data mining, pattern recognition, machine learning, etc. Manual extraction of the research papers of an individual’s interest is a time consuming and impractical task in the presence of enormous research articles. More specifically, a researcher looks for a target class guided solution, i.e., the researcher seeks for a cluster of research articles meeting his/her area of interest. To accomplish this task, this paper proposes a cluster-based parallel OCSVM \cite{scholkopf2001estimating} approach. The cluster techniques are used to group the articles properly so that the articles in the same group are more similar to each other than to those in other clusters, whereas multiple OCSVMs are trained using individual cluster information. For experiments, a set of target-tasks is defined from the domain knowledge to generalize the nature of all articles, and these tasks will become the cluster representatives. To assign the target-tasks to the clusters, OCSVM plays a decisive role. 

The clustering and classification problems are essential and admired topics of research in the area of pattern recognition and data mining. The conventional binary and multi-class classifiers are surely not suitable for this target-task mining task because, in this unsupervised learning mode, it is always possible that some of the clusters may not be assigned with any target-class, whereas an increase in the number of target-tasks to solve this problem leads to generation of duplicate information. These conventional classifiers work fine in the presence of at least two well-defined classes but may become biased, if the dataset suffers from data irregularity problems (imbalanced classes, small disjunct, skewed class distribution, missing values, etc.). Specially, when a class is ill-defined, the classifier may give biased outcome. Firstly, Minter \cite{minter1975single} observed this problem and termed it \enquote{single-class classification}, and later, Koch et al. \cite{koch1995cueing} named this fact \enquote{one-class classification} (OCC). Subsequently, researchers used different terms based on the application domain to which one-class classification was applied, like \enquote{outlier detection} \cite{ritter1997outliers}, \enquote{novelty detection} \cite{bishop1994novelty} and \enquote{concept learning} \cite{japkowicz1999concept}. In OCC problems, the target class samples are well-defined and the 
outliers are poorly defined which make decision boundary detection a difficult problem. 

For OCC, many machine-learning algorithms have been proposed like: one-class random forest (OCRF), one-class deep neural network (OCDNN), one-class nearest neighbours (OCNN), one-class support vector classifiers (OCSVCs), etc. \cite{mazhelis2006one}, \cite{pimentel2014review}, \cite{khan2014one}, \cite{chalapathy2019deep}, \cite{alam2020one}, \cite{alam2020sample}. The benifit of OCSVCs over other state-of-the-art OCC techniques is its work-ability with only positive class samples whereas the other methods need negative class samples too for smooth operation, hence the OCSVCs are found more suitable for this research. Based on extensive literature review, it is evident that OCSVMs (a type of OCSVC) are mostly used for novelty/anomaly detection in various application domains such as intrusion detection \cite{khreich2017anomaly}, \cite{maglaras2016combining}, fraud detection \cite{hejazi2013one}, \cite{dong2017method} disease diagnosis \cite{uttreshwar2008hepatitis}, \cite{cohen2008novelty} novelty detection \cite{yin2018active} and document classification \cite{manevitz2001one}. These assorted applicability make OCSVMs very interesting and important. Though many research articles have been published concerning OCSVM during the last two decades, still, it is not applied and tested for research article mining task. In this paper, OCSVM is used along with clustering techniques for article categorization, and results show that this approach gives promising performance.

The rest of the paper is organized as follows: section 2 discusses the related work in this field and details of the dataset used. The proposed model is discussed in section 3 and section 4 briefs the experimental setup and results. Finally, the conclusion is discussed in section 5.

\section{Related Work}

Document classification is an admired area of research in pattern recognition and data mining. In the present era, the presence of massive online research repositories makes the search of research articles of a user's interest, a time-consuming process. Several web search engines \cite{sanchez2018survey} are available to search keywords related documents on the web. It is observed that, still, the target class guided information retrieval from the internet is a challenging task. Specifically, for a researcher, it is a very difficult task to extract only the related articles satisfying or meeting his/her objectives. This paper proposes a novel cluster-based parallel one-class classification model to assign the most promising tasks to a group of relevant articles in available repository. The experiments have been performed with original and reduced features to justify the computational benefit of low-dimensional features. This section briefs the relevant background of data-preprocessing, clustering, one-class classification, and dimensionality reduction techniques, along with the published papers related to COVID-19 for research article mining techniques using  CORD-19 dataset.

\subsection{Data-preprocessing for text mining}
Document embeddings help to extract richer semantic content for document classification in text mining applications \cite{feldman2007text}, \cite{dai2015document}. Numeric representation of text documents is a challenging task in various applications such as document retrieval, web search, spam filtering, topic modeling, etc. In 1972, Jones \cite{robertson2004understanding} proposed a technique known as term frequency-inverse document frequency (Tf-idf). Tf-idf is a technique of information retrieval and text mining that uses a weight, which is a statistical measure to calculate the importance of a word within a document corpus. It is a frequency-based measure, and its significance increases proportionally as the frequency of the word in the corpus increases. Later, Bengio et al. \cite{bengio2003neural} proposed a feed-forward neural network language model for word embedding. Later, a simpler and more effective neural architecture was developed to learn word vectors, word2vec by Mikolov et al. \cite{mikolov2013distributed}, where the objective functions produce high-quality vectors. Recently, doc2vec as an extension of word2vec has been provided by Mikolov et al. \cite{mikolov2013efficient} to implement document-level embeddings. This technique is designed to work with the different granularity of the word sequences such as word n-gram, sentence, paragraph or document level. 
Avinash et al. \cite{joshi2020deepmine} performed intensive experiments on multiple datasets and observed that doc2vec method performs better than TF-idf. In this paper, doc2vec is used to generate features of the research documents and defined tasks. The generated features are further used in other components of the proposed model for clustering and classification.

\subsection{Clustering techniques}
Clustering is an unsupervised learning approach used iteratively to create groups of relatively similar samples from the population. In this research, following three clustering techniques have been used to create the clusters of the available samples (research articles): 
\begin{itemize}
	\item \textit{k}-means
	\item Density-based spatial clustering of applications with noise (DBSCAN)
	\item Hierarchical agglomerative clustering (HAC)
\end{itemize}

The \textit{k}-means algorithm is a popular clustering technique that aims to divide the data samples into k pre-defined distinct non-overlapping subgroups, where each data point belongs to only one group. It keeps inter-cluster data points similar to each other and also tries to maximize the distance between two clusters \cite{bradley1998refining}. DBSCAN is a data clustering algorithm proposed by Martin et al. \cite{ester1996density}. It is a non-parametric algorithm based on the concept of the nearest neighbour. It considers a set of points in some space and identifies groups of close data points as nearest neighbours and also identifies outliers to the points which are away from their neighbors. HAC \cite{mullner2011modern} is a \enquote{bottom-up} approach in which each data point is initially considered as a single-element cluster. Later, in the next steps, the two similar clusters are merged to form a bigger cluster, and subsequently, converges to a single cluster.  
Experiments have been performed on CORD-19 dataset using above mentioned clustering algorithms, and the objective is to find the best cluster representation of the whole sample space. 
\subsection{One-class classification}

One-class classification (OCC) algorithms are suitable when the negative class is either absent, poorly sampled or ill-defined. The objective of OCC is to maximize the learning ability using only the target class samples. The conventional classifiers need at least two well-defined classes for healthy operation but give biased results, if the test sample is an outlier. As a solution to this problem, the one-class classification techniques are used, majorly applicable for outliers/novelty detection and concept learning \cite{khan2009survey}. In this research, parallel one-class support vector machines are used for target-tasks assignment.

\begin{figure}
	\centering
	\includegraphics[scale=0.4] {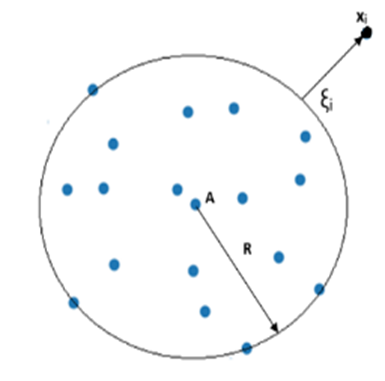}
	\caption{Support Vector Data Description (SVDD).}
	\label{fig1}
\end{figure}

Tax et al. \cite{tax1999support} solved the OCC problem by separating the target class objects from other samples in sample space, and proposed a model called support vector data description (SVDD), where target class samples are enclosed by a hypersphere, where the data points at decision boundary are treated as support vectors. The SVDD rejects a test sample as an outlier, if it falls outside of the hypersphere; otherwise accepts it as a target class sample, as shown in Fig.~\ref{fig1}. The objective function of SVDD is defined as:
\begin{equation}
\label{eq1}
\begin{aligned}
L(R,a,\alpha_i,\gamma_i,\xi_i) = R^2 + C \sum_{i} \xi_i -\sum_{i} \alpha_i \{R^2 + \xi_i \\ - (\parallel x_i\parallel^2 - 2a.x_i+ \parallel a\parallel^2)\} - \sum_{i}\gamma_i \xi_i 
\end{aligned}
\end{equation}
subject \hspace{1mm}to: \hspace{1mm}$\parallel x_i-a\parallel^2\leq R^2+\xi_i,where\hspace{2mm} \xi_i\geq 0 \hspace{2mm} \forall$ i\\
\\

where $R$ is the radius of the hypersphere (objective is to minimize the radius), data point $x_i$ is an outlier, \textit{a} is center of hypersphere, samples at decision boundary are support vectors, the parameter \textit{C} controls the trade-off between the volume and the errors, and $\xi$ is the slack variable to penalize the outlier. With the Lagrange multipliers $\alpha_i \geq 0$, $\gamma_i \geq 0$ and $i=1, 2, \dots, N$, the purpose is to minimize the hypersphere’s volume by minimizing $R$ to cover all target class samples with the penalty of slack variables for outliers. By setting partial derivatives to zero and substituting those constraints into Eq.~\ref{eq1},following is obtained:

\begin{equation}
\centering
L= \sum_{i}\alpha_i(x_i,x_i) - \sum_{i,j} \alpha_i \alpha_j (x_i, x_j)
\end{equation}

A test sample $x$ is classified as an outlier if the description value is not smaller than $C$. SVDD can also be expanded using kernels. Thus, the problem can be formulated as follows:

\begin{equation}
\centering
{\left \|  { \varphi (x) - a} \right \| }^2 \leq R^2
\end{equation}
The output of the SVDD can be calculated as follows:
\begin{equation}
\centering
\label{p}
R^2 -{\left \|  { \varphi (x) - a} \right \| }^2
\end{equation}
In Eq.~\ref{p}, the output is positive if the sample is inside the boundary, whereas for an outlier the output becomes negative.

\begin{figure}
	\centering
	\includegraphics[scale=0.3] {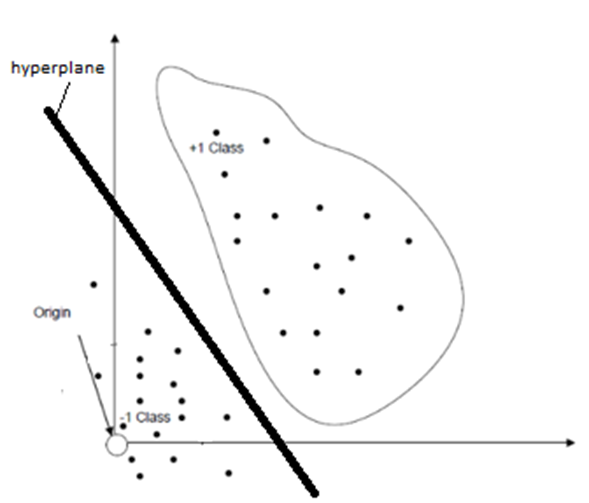}
	\caption{One-Class Support Vector Machine (OCSVM).}
	\label{fig2}
\end{figure}

Schl{\"o}kopf et al. \cite{scholkopf2001estimating} proposed an alternative approach (named one-class support vector machine (OCSVM)) to SVDD by separating the target class samples from outliers using a hyperplane instead of creating hypersphere (Fig.~\ref{fig2}). In this approach, the target class samples are separated by a hyperplane with the maximal margin from the origin and all negative class samples are assumed to fall on the subspace of the origin. This algorithm returns value +1 for target class region and -1 elsewhere. Following quadratic equation~\ref{eq2} must be solved to separate the target class samples from the origin:
\begin{equation}
\label{eq2}
\max_{w,\xi , \rho}  \frac{1}{2} \parallel w \parallel^2+ \frac{1}{\upsilon N} \sum_{i}^{n} {\xi_i - \rho }
\end{equation}
subject to: $\hspace{2mm}  w. \phi(x_i )\geq \rho-\xi_i \hspace{2mm}$ and $\hspace{2mm} \xi_i\geq 0 \hspace{2mm}$ for $\hspace{2mm}$ all  $\hspace{2mm}$ i=1, 2, 3 $\dots$ n \\
\\
where $\phi$ represents a point $x_i$ in feature space and ${\xi}_i$ is the slack variable to penalize the outlier. The objective is to find a hyperplane characterized by $\omega$ and $\rho$ to separate the target data points from the origin with maximum margin. Lower bound on the number of support vectors and upper bound on the fraction of outliers are set by $\upsilon$ $\epsilon$ (0,1]. Experimental results of this research ensures that for OCSVM, the Gaussian kernel outperforms other kernels.

The dual optimization problem of Eq.~\ref{eq2} is defined as follows:

\begin{equation}
\centering
\min_{\alpha} \frac{1}{2} \sum\limits_{i=1}^{N}\sum\limits_{j=1}^{N}{\alpha}_i{\alpha}_jK(x_i,x_j)
\end{equation}
subject to: $0 \leq{\alpha}_i \leq \frac{1}{\upsilon N}$, $\sum_{i=1}^{N}{\alpha}_i=1$,  i = 1, 2, 3 \dots, n. \\
where $\alpha = {[\alpha_1, \alpha_2, \dots, \alpha_N]}^T$ and $\alpha_i$ is the Lagrange multiplier, whereas the weight-vector $w$ can be expressed as:
\begin{equation}
\centering
w=\sum\limits_{i=0}^{N}{\alpha_i \phi x(i)}
\end{equation}
$\rho$ is the margin parameter and computed by any $x_i$ whose corresponding Lagrange multiplier satisfies $0 < {\alpha}_i < \frac{1}{\upsilon N}$

\begin{equation}
\centering
\rho= \sum\limits_{j=1}^{N}{\alpha_j K(x_j, x_i)}
\end{equation}
With kernel expansion the decision function can be defined as follows:

\begin{equation}
\centering
f(x) = \sum\limits_{i=1}^{N}{\alpha_i K(x_i, x)} - \rho
\end{equation}
Finally, the test instance $x$ can be labelled as follows:
\begin{equation}
\centering
\hat{y}= sign(f(x))
\end{equation}
where $sign(.)$ is sign function.

\begin{figure}
	\centering
	\includegraphics[scale=0.22] {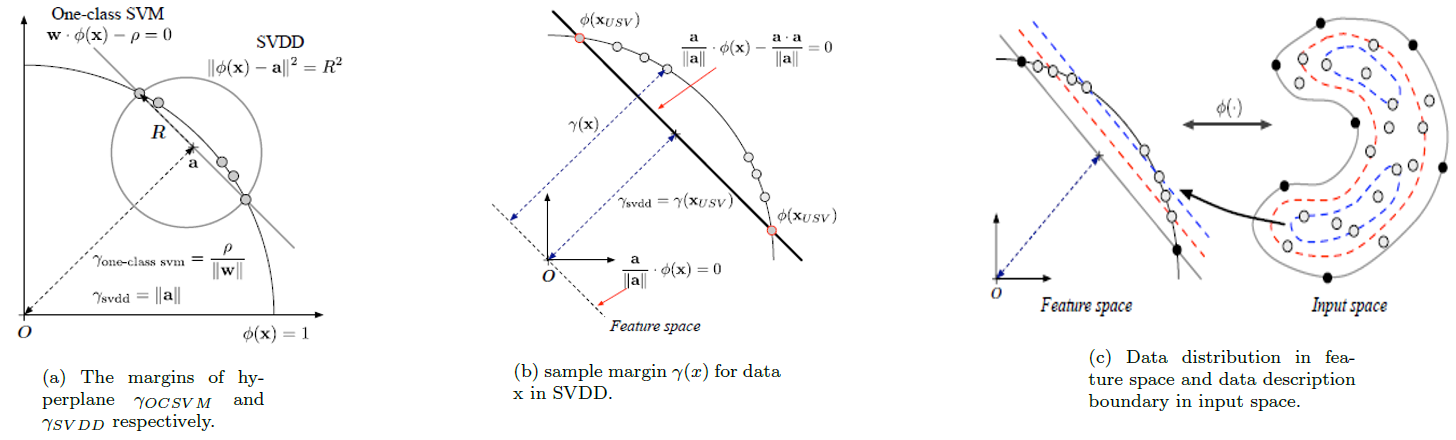}
	\caption{Geometric representation of SVDD and one-class SVM [35].}
	\label{fig12}
\end{figure}

Both these breakthrough approaches for OCC (SVDD and OCSVM) perform equally with Gaussian kernel and origin plays a decisive role where all the negative class data points are pre-assumed to lie on the origin. In unit norm feature space, the margin of a hyperplane of OCSVM is equal to the norm of the centre of SVDD is \cite{kim2008fast} as shown in Fig.~\ref{fig12}(a). SVDD can be reformulated by a hyperplane equation as follows:

\begin{equation}
\parallel \phi(x)-a\parallel^2\leq R^2 \Leftrightarrow w_{SVDD} . \phi(x)-\rho_{SVDD} \geq 0 
\end{equation}\\
where a is the centre of SVDD hypersphere. The normal vector $w_{svdd}$ and the bias $\rho_{svdd}$ of SVDD hyperplane can be defined as below: 
\begin{equation}
w_{SVDD} =\frac{a}{\parallel a \parallel}, \hspace{5mm} \rho_{SVDD} =\parallel a \parallel
\end{equation}

In feature space, the virtual hyperplane passes through the origin and the sample margin is defined by its distance from the image of the data $\emph{x}$ as shown in Fig.~\ref{fig12}(b).
The sample margin in SVDD can be defined as below:
\begin{equation}
\gamma_{SVDD}(x)=\frac{a.\phi(x)}{\parallel a \parallel}
\end{equation}\\
where $a$ is the centre of SVDD’s hypersphere and y($\emph{x}$) is the image of data $\emph{x}$ in feature space. In OCSVM, the sample margin is defined as follows:
\begin{equation}
\gamma_{OCSVM}(x)=\frac{w.\phi(x)}{\parallel w \parallel}
\end{equation}
\\
Because data examples exist on the surface of a unit hypersphere, the sample margin has \enquote*{0} as the minimum value and \enquote*{1} as the maximum, i.e.
\begin{equation}
0\leq \gamma(x) \leq 1
\end{equation}\\
Also, sample margins of unbounded support vectors  $x_{USV}$ ( 0 \textless $\alpha_{x_{USV}}$ \textless  $\frac{1}{\upsilon N}$) are the same as the margin of hyperplane. Hence :

\begin{equation}
\gamma(x_{USV})= \gamma_{SVDD}= \parallel a \parallel
\end{equation}
\begin{equation}
\gamma(x_{USV})= \gamma_{OCSVM}= \frac{\rho} {\parallel w \parallel}
\end{equation}

Sample margin represents the distribution of samples in feature space. Fig.~\ref{fig12}(c) shows the distribution of sample margin of training data and hyperplane of OCSVM in feature space. It is also evident that the SVDD and OCSVM perform equally with RBF kernel. In the present research work, OCSVM is used for experiments.

\subsection{Dimensionality reduction}

The research articles are associated with massive features and all are not equally important that leads to the curse of dimensionality. In this paper, doc2vec technique is used to generate features for experiments. Reduced features obtained by applying the following dimensionality reduction (DR) techniques are also used for experiments aiming to reduce the computation cost without affecting the performance:
\begin{itemize}
	\item Principal component analysis (PCA)
	\item Isometric Mapping (ISOMAP)
	\item t-distributed stochastic neighbor (t-SNE)
	\item Uniform manifold approximation and projection (UMAP)
\end{itemize}
Principal component analysis \cite{wold1987principal} is a linear DR technique that extracts the dominant features from the data. PCA generates a lower-dimensional representation of the data and describes the maximum variance in the data as possible. This is done by finding a linear basis of reduced dimensionality for the data, in which the amount of variance in the data is maximal. ISOMAP is a nonlinear DR technique for manifold learning that uses geodesic distances to generate features by using transformation from a larger to smaller metric space \cite{bengio2004out}. 

t-SNE works as an unsupervised, non-linear technique mainly supports data exploration and visualization in a high-dimensional space. It calculates a similarity count between pairs of instances in the high dimensional and low dimensional space and then optimizes these two similarity counts using a cost function \cite{maaten2008visualizing}. Whereas, UMAP supports effective data visualization based on Riemannian geometry and algebraic topology with excellent run time performance as compared to other techniques \cite{mcinnes2018umap}.

\subsection{Recent researches on CORD-19 dataset}

The COVID-19 open research dataset (CORD-19) \cite{kohlmeier2020covid} is prepared by White House in partnership with leading research groups characterizing the wide range of literature related to coronavirus. It consists of 45,000 scholarly articles, within 33,000 articles are full text about various categories of coronavirus. The dataset is a collection of commercial, noncommercial, custom licensed and bioRxiv/medRxiv subsets of documents from multiple repositories contributed by various research groups all over the world.  The CORD-19 dataset is hosted by kaggle, consists of a set of useful questions about disease spread in order to find out information regarding its origin, causes, transmission, diagnostic, etc. These questions can motivate researchers to explore preceding epidemiological studies to have better planning of preventive measures under present circumstances of COVID-19 disease outbreak. This dataset is aimed to cater the global research community, an opportunity to perform machine learning, data mining and natural language processing tasks to explore the hidden insights within it and utilize the knowledge to tackle this pandemic worldwide. Thus, there is an utmost need to explore the literature with minimum time and effort so that all possible solutions related to the worldwide pandemic could be achieved.

In a research initiative, Wang et al. \cite{Wang2020ss} generated a named entity recognition (NER) based dataset from CORD-19 corpus (2020-03-13). This derived CORD-19-NER dataset consists of 75 categories of entities extracted from 4 different data sources. The data generation method is a weakly supervised method and useful for text mining in both biomedical as well as societal applications. Later, Han et al. \cite{han2020effects} in their research work, focused on the impact of outdoor air pollution on the immune system. The research work highlighted that increased air pollution causes respiratory virus infection, but during the lock-down, along with social distancing and home isolation measures, air pollution gets reduced. This research work utilizes daily confirmed COVID-19 cases in selected cities of China, air pollution, meteorology data, intra/inter city level movements, etc. They proposed a regression model to establish the relationship between the infection rate and other surrounding factors. 

Since the entire world is suffering from the severity of the novel coronavirus outbreak,  researchers are keen to learn more and more about coronavirus. In a research work, Dong et al. \cite{dong2020understand} proposed a latent Dirichlet allocation (LDA) based topic modeling approach using CORD-19 dataset. This article highlights the areas with limited research; therefore, future research activities can be planned accordingly. The corona outbreak pandemic situation demands extensive research on the corona vaccine so that we can control this outbreak. Manual methods of exploring the available literature are time-consuming; therefore, Joshi et al. \cite{joshi2020deepmine}, proposed deep learning-based automatic literature mining for summarization of research articles for faster access.

It is observed that mining scholarly articles is a promising area of research when enormous papers are available concerning COVID-19. Search engines may give unrelated articles, and therefore the overall search process becomes ineffective and time-consuming. As the overall analysis is supposed to perform in unlabeled data, the clustering technique helps to group the related articles in an optimized way. It is obvious that the search query is always task specific, hence the target-tasks are defined with domain knowledge. Conventional classifiers (binary or multi-class) may give biased results and never ensure task-specific classification, thus parallel one-class SVMs are used for target-task assignment to clusters. This method helps the researchers to instantly find the desired research articles more precisely. For experiments, the CORD-19 dataset is used to validate the performance of the proposed approach.

\section{Proposed Methodology}

\begin{figure}
	\centering
	\includegraphics[scale=0.22] {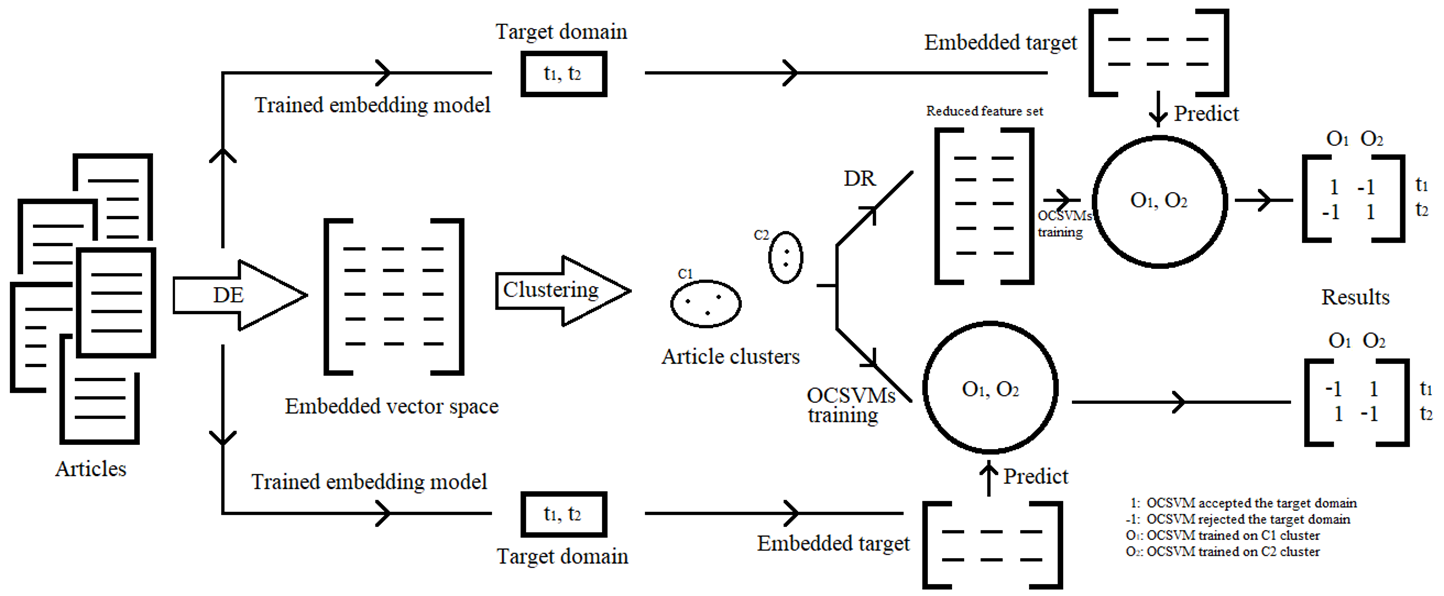}
	\caption{Schematic representation of the proposed approach.}
	\label{fig3}
\end{figure}

This research work offers a bottom-up approach to mine the COVID-19 articles concerning the target-tasks. These tasks are defined and prepared by domain knowledge to answer all possible queries of coronavirus related research articles. In this paper, instead of associating the articles to address the target-tasks, the intention is to map the defined tasks to the cluster of similar articles. The overall approach is divided into several components, such as document embedding (DE), articles clustering (AC), dimensionality reduction (DR) and visualization, and one-class classifier (OCC) as shown in Fig.~\ref{fig3}

Since the abstract of a scholarly article represents the semantic meaning of the whole article, this along with the target-tasks are mapped onto the numeric data represented in multi-dimensional vector space using the state-of-the-art natural language processing (NLP) document embedding techniques such as doc2vec \cite{lau2016empirical} based on distributed memory version of paragraph vector (PV-DM) \cite{le2014distributed}. The generated vector spaces of the article's abstracts are utilized to generate segregated clusters of related articles. The extensive trials are conducted with varying clustering approaches like \textit{k}-means \cite{bradley1998refining}, DBSCAN \cite{ester1996density} and HAC \cite{mullner2011modern}, to generate an appropriate number of segregated clusters. The generated clusters are also analyzed with the two-dimensional visual representation via dimensionality reduction techniques such as PCA \cite{wold1987principal}, ISO- MAP \cite{bengio2004out}, t-SNE \cite{maaten2008visualizing}, and UMAP \cite{mcinnes2018umap}.  Each of the generated clusters are parallely trained on one-class support vector machines (OCSVM) separately. The trained models are then utilized to associate the most appropriate articles for the concerned required information (target-task). Furthermore, these generated results are statistically compared in contrast to the computational effective approach (CEA), where in CEA, the OCSVM models are trained on reduced clusters feature set generated using above discussed dimensionality reduction techniques. Algorithm 1 presents the complete process of the proposed approach.\\
\\ \textbf{Algorithm:}  Cluster creation and target-task assignment \\
\\\textbf{Input:} \\ Dataset $ = {A_1, A_2, \dots, A_d}$         //${A_i}^s$ are article  \\
\textbf{Initialization:} \\Cluster$\_$Tech [ ]$ =  [\textit{k}-means, DBSCAN, HAC ]$, \\Target$\_$Task[ ]$ =[T_1, T_2, \dots, T_N]$ \\
\\\textbf{Step 1:} Repeat Step 2, 3 ,4 and 5 for every cluster technique  in Cluster$\_$Tech    \\
\phantom{Step 1:}\hspace{4ex} Create$\_$clusters (Cluster$\_$Tech [ i ], Dataset)  \\
\phantom{Step 1:}\hspace{4ex} Return ( Clusters [ i ], ${\#}$Clusters [i] )  \\ \phantom{Step 1:}\hspace{10ex}//$\#$Clusters is Optimum number of clusters  \\
\\\textbf{Step 2:}  Train parallel one class SVMs for all the clusters generated by a \\ \phantom{Step 2:}\hspace{3ex} given technique \\
\phantom{Step 2:}\hspace{4ex}   OCSVM (Clusters [ i ]) \\
\phantom{Step 2:}\hspace{4ex}	//OCSVM($C_1$) $\parallel$ \dots $\parallel$ OCSVM($C_{\#Clusters}$)\\
\\\textbf{Step 3:} //Validation and Testing the models \\
\phantom{Step 3:}\hspace{4ex}  For $\#$Clusters in Cluster [ i ] \\
\phantom{Step 3:}\hspace{4ex} Test$\_$OCSVM( j ) $<-$ (+ve Samples, Outliers)  \\ \phantom{Step 3:}\hspace{4ex} //j= $ {C_1, C_2, \dots, C_{\#Clusters}}$ \\
\\\textbf{Step 4:} // Assignment of target-tasks  \\
\phantom{Step 4:}\hspace{4ex} Repeat for All defined Tasks (N) \\
\phantom{Step 4:}\hspace{4ex} Test$\_$OCSVM( j ) $<-$ ($T_j$)\\
\\\textbf{Step 5:} Return (OCSVMs, Target$\_$IDs)\\
\\\textbf{Step 6:} Repeat Steps 1 to 5 with reduced features.\\
\\\textbf{Step 7:} Stop.

\section{Experimentation}
The proposed methodology is trained and evaluated on CORD-19 \cite{kohlmeier2020covid} dataset, retrieved on April 4, 2020. After extensive study, the articles and the target domain comprising nine target-tasks as shown in Table~\ref{tab1} \cite{kohlmeier2020covid}. The details of these tasks (not included in this article because of length) are later explored to generate features for further processing. These task descriptions are embedded in a high dimensional vector space using doc2vec approach. After extensive trials it is evident that 150 features are sufficient for optimized performance.
\begin{table}[]
	\centering
	\caption{Target description from CORD-19 \cite{kohlmeier2020covid}.}
	\label{tab1}
	\begin{adjustbox}{width=\columnwidth,center}{
	\begin{tabular}{|p{0.7cm}|p{7cm}|}
		\hline
		Target ID & Target Domain                        \\ \hline
		T-1   & What is known about transmission incubation and environmental stability?          \\ \hline
		T-2   & What do we know about COVID-19 risk factors? What have we learned from epidemiological studies?  
		\\ \hline
		T-3       &  What do we know about virus genetics origin and evolution?"                        
		\\ \hline
		T-4       & What do we know about vaccines and therapeutics? What has been published concerning research and development and evaluation efforts of vaccines and therapeutics? \\ \hline
		T-5       & What do we know about the effectiveness of non-pharmaceutical interventions?                                         \\ \hline
		T-6       & What do we know about diagnostics and surveillance?                                      \\ \hline
		T-7       & What has been published about medical care?                                              \\ \hline
		T-8       & What has been published concerning ethical considerations for research?                 \\ \hline
		T-9       &  What has been published about information sharing and inter-sectoral collaboration?  \\ \hline                                                                        
	\end{tabular}}
	\end{adjustbox}
	\end{table}

The doc2vec implementation is based on the PV-DM model that is analogous to the continuous bag of words (CBOW) approach of word2vec \cite{le2014distributed} with just an additional feature vector representing the full article. The PV-DM model obtains the document vector by training a neural network to predict the words from the context of vocabulary or word vector and full doc-vector, represented as \textit{W} and \textit{D} \cite{le2014distributed}. Following this, the word model is trained on words $w_1$, $w_2$, \dots, $w_T$, with the objective to maximize the log likelihood as given in Eq~\ref{eq3}. The word is predicted with probability distribution obtained using softmax activation function provided in Eq~\ref{eq4}. The Eq~\ref{eq5} indicates the hypothesis ($y_i$) associated with each output word ($i$).
\begin{equation}
\label{eq3}
\frac{1}{T}\sum\limits_{t=k}^{T-k}{\log p(w_t|w_{t-k}....w_{t+k})}
\end{equation}
where $\textit{k}$ indicates the window size to consider the words context.

\begin{equation}
\label{eq4}
p(w_t|w_{t-k}....w_{t+k})= \frac{e^{y_{w_t}}}{\sum\limits_{i}^{}{e^{y_{i}}}} \end{equation}	

\begin{equation}
\label{eq5}
y=b+Uh(W,D)
\end{equation}	
where $\textit{U}$ and $\textit{b}$ are trainable parameters and $h$ is obtained by concatenation of word vectors ($W$) and document vector ($D$).

This approach is utilized to embed the target-tasks and corpus of COVID-19 articles into high dimensional vector space. Due to the small corpus of targets, instead of directly generating the target vector space, each target query is elaborated to emphasize on the expected type of articles by adding the appropriate semantic meaning.

\begin{figure}
	\centering
	\includegraphics[scale=0.2] {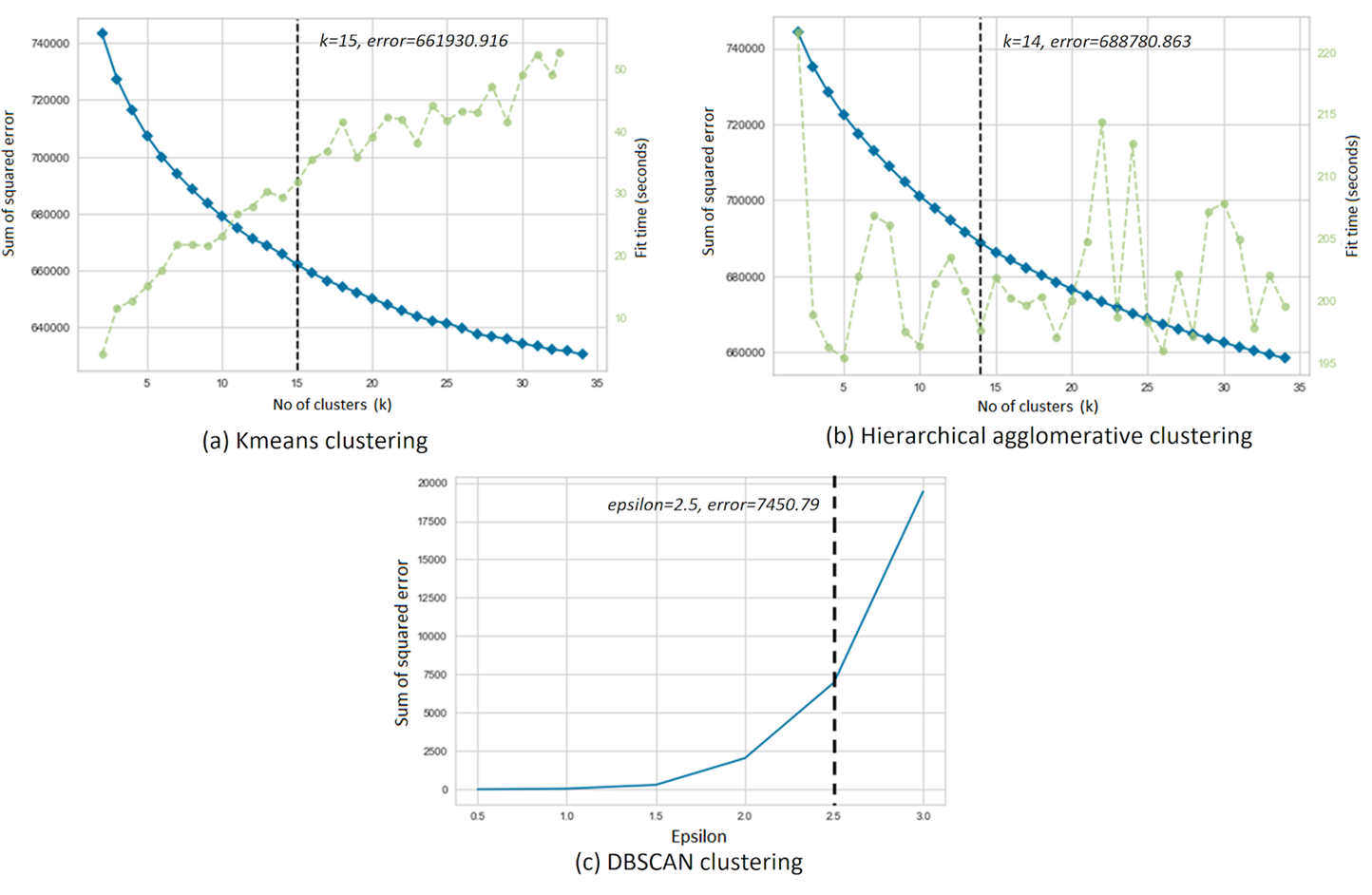}
	\caption{Computing most suitable number of clusters.}
	\label{fig4}
\end{figure}

The generated document vector space of the articles is utilized by clustering approaches; \textit{k}-means, DBSCAN, and HAC. In \textit{k}-means and HAC, the appropri- ate number of clusters are set by computing the sum of squared error for the number of clusters varying from 2 to 35. As shown in Fig.~\ref{fig4}(a) and~\ref{fig4}(b), the elbow point indicates that there are 14 and 15 favourable numbers of clusters that can represent the most related articles. Whereas DBSCAN approach considers groups of close data points (minimum sample points) in a defined space (epsilon) as nearest neighbours without specifying the number of clusters. Fig.~\ref{fig4}(c) illustrates the sum of squared error for the generated clusters with varying epsilon value. The epsilon is kept as 2.5 due to the sudden increase in the sum of squared error for higher values which resulted in skewed 16 clusters of articles. 

\begin{figure}
	\centering
	\includegraphics[width=\linewidth] {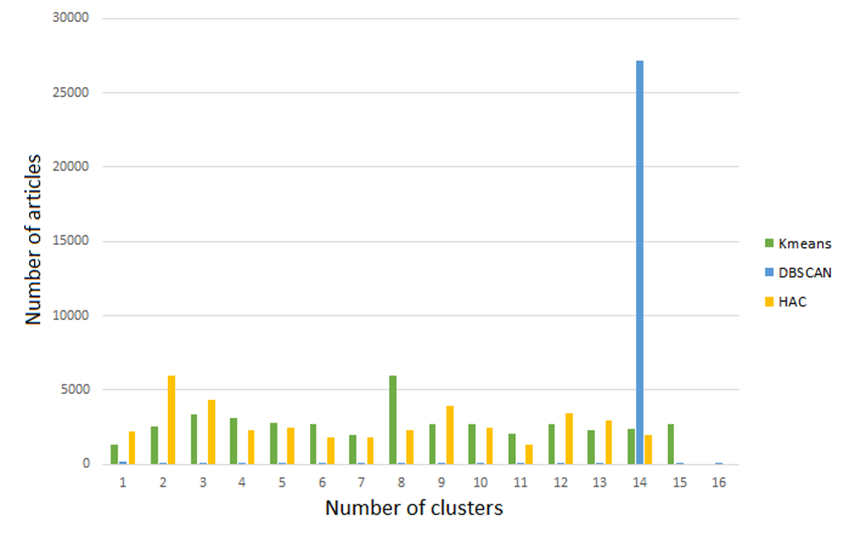}
	\caption{Distribution of samples in each cluster formed by using \textit{k}-means, DBSCAN and HAC.}
	\label{fig5}
\end{figure}

Fig.~\ref{fig5} presents the number of articles constituting each cluster generated from the
above discussed clustering algorithm out of which DBSCAN performs the worst due to the sparse mapping of the articles in the cluster.

\begin{table*}[]
	\centering
	\caption{Prediction analysis of the OCSVMs trained on \textit{k}-means clusters corresponding  to each target from CORD-19 dataset.}
	\label{tab2}
\begin{adjustbox}{width=\textwidth,center}{	\begin{tabular}{|p{.8cm}|p{15cm}|p{1cm}|p{1cm}|p{1cm}|}
		\hline
		Target ID & Most relevant article   & Similarity score & Associated cluster & Total articles \\ \hline
		\multirow{5}{*}{T-1}   & Persistence of Antibodies against Middle East Respiratory Syndrome coronavirus     & 0.4                      & 1         & \multirow{5}{*}{7953}               \\ \cline{2-4} & Aerodynamic Characteristics and RNA Concentration of SARS-CoV-2 Aerosol in Wuhan  Hospitals during COVID-19 Outbreak  & 0.4              & 12  &    \\ \cline{2-4}
		& Evaluation of SARS-CoV-2 RNA shedding in clinical specimens and clinical characteristics of 10 patients with COVID-19 in Macau            & 0.39   & 8                                       &     \\ \cline{2-4}  & Effects of temperature on COVID-19 transmission             & 0.37     & 12                 &     \\ \cline{2-4}  & Human coronavirus 229E Remains Infectious on Common Touch Surface Materials         & 0.37             & 11       &    \\ \hline
		\multirow{5}{*}{T-2}            & Clinical and Epidemiologic Characteristics of Spreaders of Middle East Respiratory Syndrome coronavirus during the 2015 Outbreak in Korea                                                               & 0.51                                  & 12                                      & \multirow{5}{*}{20630}              \\ \cline{2-4}
		& Comparative Pathogenesis of Covid-19, Mers And Sars In A Non-Human Primate Model                                                                                                                                                                    & 0.51                                  & 12                                      &                                     \\ \cline{2-4}
		& A comparison study of SARS-CoV-2 IgG antibody between male and female COVID-19  patients: a possible reason underlying different outcome between gender                                               & 0.49                                  & 5                                       &                                     \\ \cline{2-4}
		& Severe acute respiratory syndrome (SARS) in intensive care units (ICUs): limiting the risk  to healthcare workers                                                                                        & 0.48                                  & 5                                       &                                     \\ \cline{2-4}
		& Comparison of viral infection in healthcare- associated pneumonia (HCAP) and  community-acquired pneumonia (CAP)                                                                                        & 0.48                                  & 5                                       &                                     \\ \hline
		\multirow{5}{*}{T-3}            & The impact of within-herd genetic variation upon inferred transmission trees for foot-and-mouth disease virus                                                                                           & 0.43                                  & 3                                       & \multirow{5}{*}{9412}               \\ \cline{2-4}
		& Epidemiologic data and pathogen genome sequences: a powerful synergy for public health                                                                                                                                                              & 0.42                                  & 2                                       &                                     \\ \cline{2-4}
		& Transmission Parameters of the 2001 Foot and Mouth Epidemic in Great Britain                                                                                                                                                                        & 0.39                                  & 3                                       &                                     \\ \cline{2-4}
		& Middle East respiratory syndrome coronavirus neutralising serum antibodies in dromedary camels: a comparative serological study                                                                         & 0.39                                  & 11                                      &                                     \\ \cline{2-4}
		& FastViromeExplorer: a pipeline for virus and phage identification and abundance profiling in metagenomics data                                                                                          & 0.39                                  & 13                                      &                                     \\ \hline
		\multirow{5}{*}{T-4}            & Development of a recombinant truncated nucleocapsid protein based immunoassay for detection of antibodies against human coronavirus OC43                                                                 & 0.42                                  & 10                                      & \multirow{5}{*}{13499}              \\ \cline{2-4}
		& The Effectiveness of Convalescent Plasma and Hyperimmune Immunoglobulin for the Treatment of Severe Acute Respiratory Infections of Viral Etiology: A Systematic Review and Exploratory Meta-analysis & 0.4                                   & 3                                       &                                     \\ \cline{2-4}
		& Journal Pre-proof Discovery and development of safe-in-man broad-spectrum antiviral agents Discovery and Development of Safe-in-man Broad-Spectrum Antiviral Agents                                      & 0.4                                   & 10                                      &                                     \\ \cline{2-4}
		& Evaluation of Group Testing for SARS-CoV-2 RNA                                                                                                                                                                                                      & 0.4                                   & 3                                       &                                     \\ \cline{2-4}
		& Antiviral Drugs Specific for coronaviruses in Preclinical Development                                                                                                                                                                               & 0.39                                  & 10                                      &                                     \\ \hline
		\multirow{5}{*}{T-5}            & Communication of bed allocation decisions in a critical care unit and accountability for reasonableness                                                                                                 & 0.45                                  & 2                                       & \multirow{5}{*}{12170}              \\ \cline{2-4}
		& Collaborative accountability for sustainable public health: A Korean perspective on the effective use of ICT-based health risk communication                                                             & 0.44                                  & 7                                       &                                     \\ \cline{2-4}
		& Pandemic influenza control in Europe and the constraints resulting from incoherent public health laws                                                                                                    & 0.44                                  & 11                                      &                                     \\ \cline{2-4}
		& Street-level diplomacy and local enforcement for meat safety in northern Tanzania: knowledge, pragmatism and trust                                                                                       & 0.44                                  & 2                                       &                                     \\ \cline{2-4}
		& Staff perception and institutional reporting: two views of infection control compliance  in British Columbia and Ontario three years after an outbreak of severe acute respiratory syndrome            & 0.43                                  & 2                                       &                                     \\ \hline
		\multirow{5}{*}{T-6}            & Planning and preparing for public health threats at airports                                                                                                                                                                                        & 0.42                                  & 10                                      & \multirow{5}{*}{8777}               \\ \cline{2-4}
		& Can free open access resources strengthen knowledge-based emerging public health priorities, policies and programs in Africa?                                                                            & 0.4                                   & 10                                      &                                     \\ \cline{2-4}
		& Implications of the One Health Paradigm for Clinical Microbiology                                                                                                                                                                                   & 0.39                                  & 2                                       &                                     \\ \cline{2-4}
		& Annals of Clinical Microbiology and Antimicrobials Predicting the sensitivity and specificity of published real-time PCR assays                                                                         & 0.38                                  & 3                                       &                                     \\ \cline{2-4}
		& A field-deployable insulated isothermal RT-PCR assay for identification of influenza A (H7N9) shows good performance in the laboratory                                                                  & 0.38                                  & 7                                       &                                     \\ \hline
		\multirow{5}{*}{T-7}            & Strategy and technology to prevent hospital-acquired infections: Lessons from SARS, Ebola, and MERS in Asia and West Africa                                                                              & 0.41                                  & 15                                      & \multirow{5}{*}{9912}               \\ \cline{2-4}
		& Rapid Review The psychological impact of quarantine and how to reduce it: rapid review of the evidence                                                                                                 & 0.41                                  & 4                                       &                                     \\ \cline{2-4}
		& Factors Informing Outcomes for Older Cats and Dogs in Animal Shelters                                                                                                                                                                               & 0.4                                   & 4                                       &                                     \\ \cline{2-4}
		& A Critical Care and Transplantation-Based Approach to Acute Respiratory Failure after Hematopoietic Stem Cell Transplantation in Children                                                              & 0.39                                  & 8                                       &                                     \\ \cline{2-4}
		& Adaptive multiresolution method for MAP reconstruction in electron tomography                                                                                                                                                                       & 0.39                                  & 14                                      &                                     \\ \hline
		\multirow{5}{*}{T-8}            & Twentieth anniversary of the European Union health mandate: taking stock of perceived achievements, failures and missed opportunities - a qualitative study                                            & 0.47                                  & 12                                      & \multirow{5}{*}{15191}              \\ \cline{2-4}
		& Towards evidence-based, GIS-driven national spatial health information infrastructure and surveillance services in the United Kingdom                                                                    & 0.46                                  & 7                                       &                                     \\ \cline{2-4}
		& H1N1 influenza pandemic in Italy revisited: has the willingness to get vaccinated suffered in the long run?                                                                                              & 0.44                                  & 7                                       &                                     \\ \cline{2-4}
		& What makes health systems resilient against infectious disease outbreaks and natural hazards? Results from a scoping review                                                                              & 0.44                                  & 14                                      &                                     \\ \cline{2-4}
		& A qualitative study of zoonotic risk factors among rural communities in southern China                                                                                                                                                              & 0.44                                  & 14                                      &                                     \\ \hline
		
		\multirow{5}{*}{T-9}            & A cross-sectional study of pandemic influenza health literacy and the effect of a public health campaign                                                                                                & 0.5                                   & 1                                       & \multirow{5}{*}{12332}              \\ \cline{2-4}
		& Public Response to Community Mitigation Measures for Pandemic Influenza                                                                                                                                                                             & 0.48                                  & 1                                       &                                     \\ \cline{2-4}
		& Casualties of war: the infection control assessment of civilians transferred from conflict zones to specialist units overseas for treatment                                                              & 0.47                                  & 8                                       &                                     \\ \cline{2-4}
		& Communications in Public Health Emergency Preparedness: A Systematic Review of the Literature                                                                                                                                                       & 0.47                                  & 4                                       &                                     \\ \cline{2-4}
		& Ethics-sensitivity of the Ghana national integrated strategic response plan for pandemic influenza                                                                                                                                                  & 0.46                                  & 4                                       &                                     \\ \hline
	\end{tabular}}
	\end{adjustbox}
\end{table*}

\begin{figure*}
	\centering
	\includegraphics[width=0.8 \linewidth] {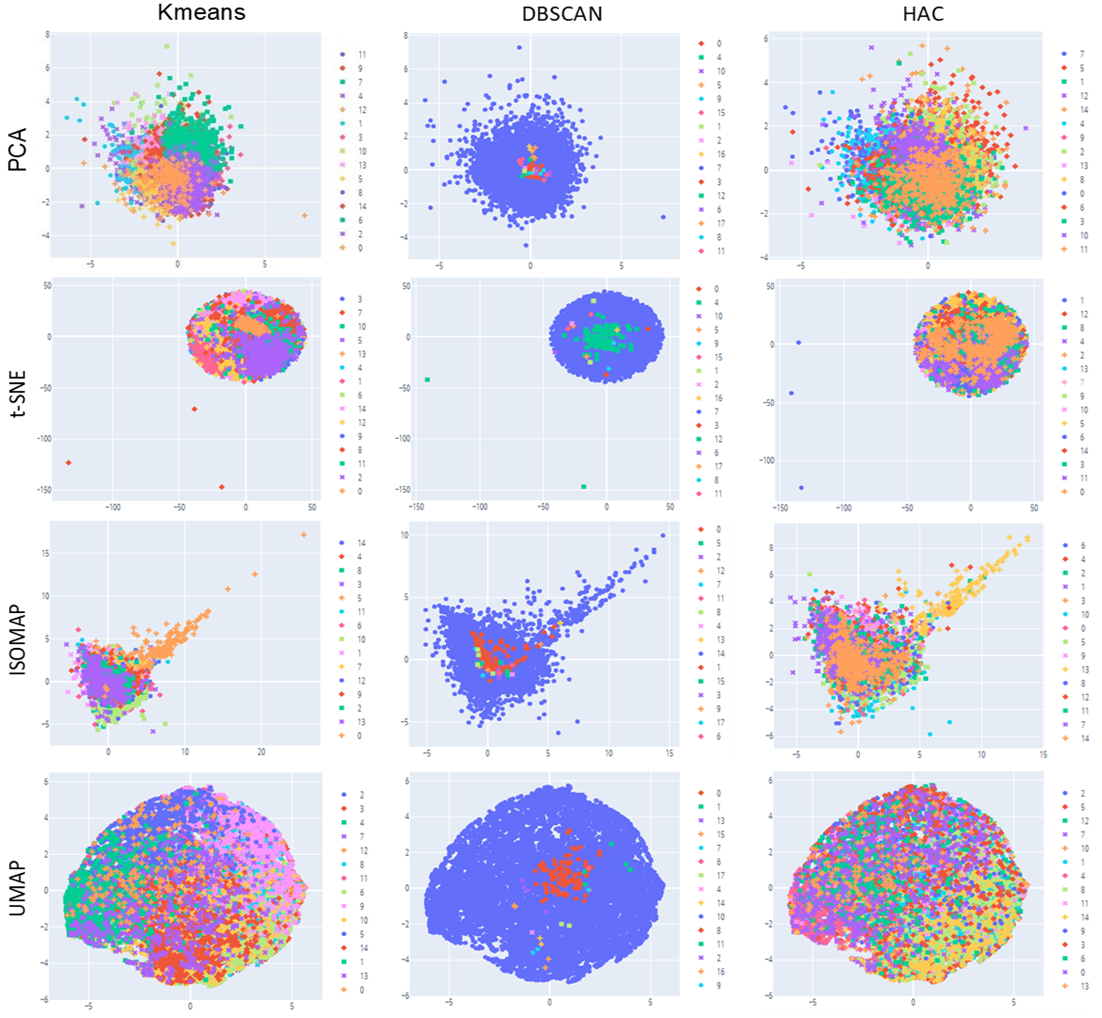}
	\caption{Clustering visualization of corpus of COVID-19 articles. }
	\label{fig6}
\end{figure*}

\begin{figure*}
	\centering
	\includegraphics[width=0.8 \linewidth] {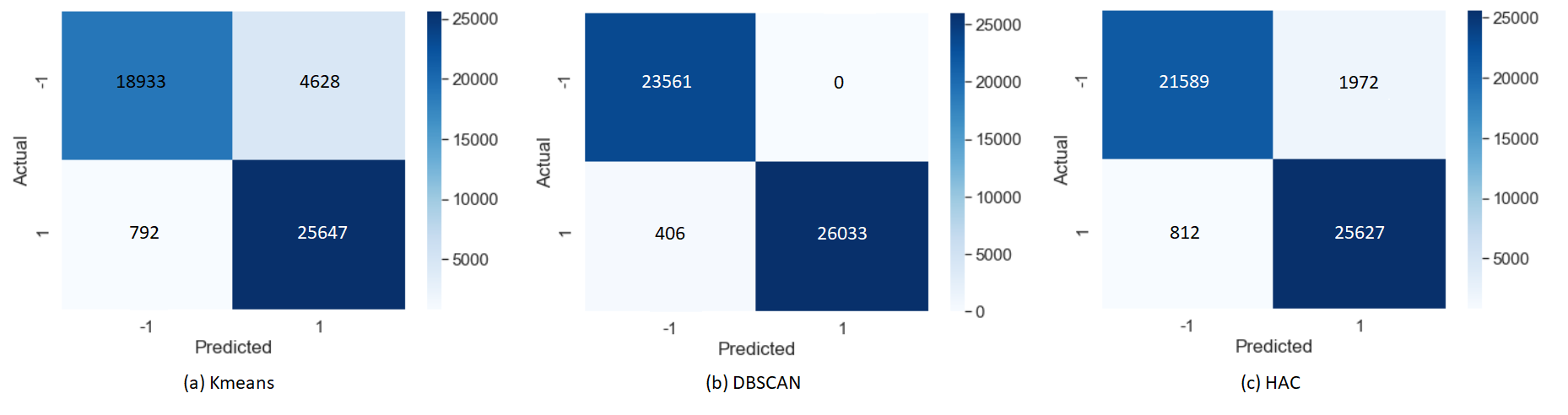}
	\caption{Normalized confusion matrix for average performance of the parallel OCSVMs.}
	\label{fig7}
\end{figure*}

Since these clusters are featured in high dimensional space, visualizing them in two dimensional space requires dimensionality reduction techniques like PCA, ISOMAP, etc. Later, these reduced feature-sets are also utilized to train the OCSVMs in order to compare the results in presence of original feature space. The clusters are visualized with the principal components displaying highest variance. Fig.~\ref{fig6} illustrates the two dimensional cluster representation from the multi-dimensional feature vector of subsample of articles for each possible combination of the discussed clustering and dimensionality reduction approaches. It is evident that from the visualization shown in Fig.~\ref{fig6}, among the utilized clustering approaches, \textit{k}-means is able to generate the meaningful clusters with better segregation and aggregation of COVID-19 articles.

There are 14, 15 and 16 numbers of clusters generated from HAC, \textit{k}-means, and DBSCAN respectively. For each of the generated clusters, dedicated OCSVMs are trained parallely where each OCSVM learns to confine a cluster on the positive side of the hyperplane with a maximum margin from the origin. The trained models are utilized to predict the most related target queries concerning a  cluster as per Eq.~\ref{eq6}. 

\begin{equation}
O_t(T_t)=\begin{cases}
+1, & if C_i \epsilon T_t .\\
-1, & \text{otherwise}.
\end{cases}		
\label{eq6}
\end{equation}

\begin{equation}
S(P,T)= \frac{P.T}{{\left \|{P} \right \|}. {\left \|{T} \right \|}}	
\label{eq7}
\end{equation}		
where $P$ and $T$ indicate any vector space.
where $O_i$ indicates the $i^{th}$ OCSVM trained on $C_i$ cluster, and $T_t$ indicates the target-task for which to identify the related articles. Each predicted target domain is verified using the cosine similarity metric as given in Eq.~\ref{eq7} in contrast to the assigned clusters of articles. The metric value ranges between 0 and 1, with the meaning of articles being totally different and same respectively. Finally, the articles are sorted in the order of most relevance based on the highest cosine score. The Table~\ref{tab2} presents the top five related articles and the corresponding similarity score along with the total number of articles found with the cosine score greater than 0.1, using the OCSVMs trained on the clusters generated via \textit{k}-means approach. It is also observed that the inter-cluster similarity is always less that 0.1 that plays a significant role in target class classification of COVID-19 based research articles.

\begin{table*}[]
	\centering
	\caption{Target-task mapping using OCSVMs end-to-end trained on article clusters.}
	\label{tab3}
	\begin{adjustbox}{width=0.8 \textwidth,center}{	\begin{tabular}{|l|l|l|l|l|l|l|l|l|l|l|l|l|l|l|l|l|l|}
				\hline
				\multirow{2}{*}{CA}      & \multirow{2}{*}{Tasks} & \multicolumn{16}{l|}{OCSVMs trained on clusters}                                                            \\ \cline{3-18} 
				&                        & 1    & 2    & 3    & 4    & 5    & 6    & 7    & 8    & 9    & 10   & 11   & 12   & 13   & 14   & 15   & 16 \\ \hline
				\multirow{9}{*}{\textit{k}-means} & T1                     & 1    & -1   & -1   & 1    & -1   & -1   & -1   & 1    & -1   & -1   & 1    & 1    & -1   & -1   & -1   & -  \\ \cline{2-18} 
				& T2                     & 1    & -1   & -1   & -1   & 1    & 1    & 1    & 1    & -1   & 1    & 1    & 1    & -1   & 1    & 1    & -  \\ \cline{2-18} 
				& T3                     & -1   & 1    & 1    & -1   & -1   & -1   & -1   & -1   & -1   & -1   & 1    & 1    & -1   & -1   & -1   & -  \\ \cline{2-18} 
				& T4                     & -1   & -1   & 1    & -1   & -1   & -1   & -1   & -1   & -1   & 1    & -1   & -1   & 1    & -1   & -1   & -  \\ \cline{2-18} 
				& T5                     & -1   & -1   & 1    & -1   & -1   & -1   & 1    & -1   & -1   & -1   & 1    & 1    & -1   & -1   & -1   & -  \\ \cline{2-18} 
				& T6                     & -1   & 1    & 1    & -1   & -1   & -1   & 1    & -1   & -1   & 1    & -1   & -1   & 1    & -1   & 1    & -  \\ \cline{2-18} 
				& T7                     & 1    & -1   & -1   & 1    & -1   & -1   & -1   & -1   & -1   & 1    & -1   & -1   & 1    & -1   & 1    & -  \\ \cline{2-18} 
				& T8                     & 1    & 1    & -1   & -1   & -1   & 1    & 1    & 1    & 1    & -1   & 1    & 1    & -1   & 1    & 1    & -  \\ \cline{2-18} 
				& T9                     & 1    & -1   & -1   & 1    & -1   & -1   & -1   & 1    & -1   & -1   & -1   & -1   & 1    & -1   & -1   & -  \\ \hline
				\multicolumn{2}{|l|}{\% score}                    & 0.44 & 0.56 & 0.56 & 0.67 & 0.89 & 0.78 & 0.56 & 0.44 & 0.89 & 0.56 & 0.44 & 0.44 & 0.56 & 0.78 & 0.56 & -  \\ \hline
				\multirow{9}{*}{DBSCAN}  & T1                     & -1   & -1   & -1   & -1   & -1   & -1   & -1   & -1   & -1   & -1   & -1   & -1   & -1   & 1   & -1   & -1 \\ \cline{2-18} 
				& T2                     & -1   & -1   & -1   & -1   & -1   & -1   & -1   & -1   & -1   & -1   & -1   & -1   & -1   & 1   & -1   & -1 \\ \cline{2-18} 
				& T3                     & -1   & -1   & -1   & -1   & -1   & -1   & -1   & -1   & -1   & -1   & -1   & -1   & -1   & 1   & -1   & -1 \\ \cline{2-18} 
				& T4                     & -1   & -1   & -1   & -1   & -1   & -1   & -1   & -1   & -1   & -1   & -1   & -1   & -1   & 1   & -1   & -1 \\ \cline{2-18} 
				& T5                     & -1   & -1   & -1   & -1   & -1   & -1   & -1   & -1   & -1   & -1   & -1   & -1   & -1   & 1   & -1   & -1 \\ \cline{2-18} 
				& T6                     & -1   & -1   & -1   & -1   & -1   & -1   & -1   & -1   & -1   & -1   & -1   & -1   & -1   & 1   & -1   & -1 \\ \cline{2-18} 
				& T7                     & -1   & -1   & -1   & -1   & -1   & -1   & -1   & -1   & -1   & -1   & -1   & -1   & -1   & 1   & -1   & -1 \\ \cline{2-18} 
				& T8                     & -1   & -1   & -1   & -1   & -1   & -1   & -1   & -1   & -1   & -1   & -1   & -1   & -1   & 1   & -1   & -1 \\ \cline{2-18} 
				& T9                     & -1   & -1   & -1   & -1   & -1   & -1   & -1   & -1   & -1   & -1   & -1   & -1   & -1   & 1   & -1   & -1 \\ \hline
				\multicolumn{2}{|l|}{\% score}                    & 0    & 0    & 0    & 0    & 0    & 0    & 0    & 0    & 0    & 0    & 0    & 0    & 0    & 0    & 0    & 0  \\ \hline
				\multirow{9}{*}{HAC}     & T1                     & 1    & -1   & -1   & 1    & -1   & -1   & -1   & 1    & -1   & -1   & 1    & 1    & -1   & -1   & -    & -  \\ \cline{2-18} 
				& T2                     & 1    & -1   & -1   & -1   & 1    & 1    & 1    & 1    & -1   & 1    & 1    & 1    & -1   & 1    & -    & -  \\ \cline{2-18} 
				& T3                     & -1   & 1    & 1    & -1   & -1   & -1   & -1   & -1   & 1    & -1   & 1    & 1    & -1   & -1   & -    & -  \\ \cline{2-18} 
				& T4                     & -1   & -1   & 1    & -1   & 1    & -1   & -1   & -1   & -1   & 1    & -1   & -1   & 1    & -1   & -    & -  \\ \cline{2-18} 
				& T5                     & -1   & 1    & 1    & -1   & -1   & -1   & 1    & -1   & -1   & -1   & 1    & 1    & -1   & -1   & -    & -  \\ \cline{2-18} 
				& T6                     & -1   & 1    & 1    & -1   & -1   & -1   & 1    & -1   & 1    & 1    & -1   & -1   & 1    & -1   & -    & -  \\ \cline{2-18} 
				& T7                     & -1   & -1   & -1   & 1    & -1   & 1    & -1   & 1    & -1   & 1    & -1   & -1   & 1    & -1   & -    & -  \\ \cline{2-18} 
				& T8                     & 1    & 1    & -1   & -1   & -1   & 1    & 1    & 1    & 1    & -1   & 1    & 1    & -1   & 1    & -    & -  \\ \cline{2-18} 
				& T9                     & 1    & -1   & 1    & 1    & -1   & -1   & -1   & 1    & -1   & 1    & -1   & -1   & 1    & -1   & -    & -  \\ \hline
				\multicolumn{2}{|l|}{\% score}                    & 0.56 & 0.56 & 0.44 & 0.67 & 0.78 & 0.67 & 0.56 & 0.44 & 0.67 & 0.44 & 0.44 & 0.44 & 0.56 & 0.78 & -    & -  \\ \hline
			\end{tabular}}
		\end{adjustbox}
	\end{table*}

\section{Results and discussion}

The Table~\ref{tab3} illustrates the detailed results of the OCSVMs to map the target domain to the group of articles trained on each cluster generated using \textit{k}-means, DBSCAN, and HAC. The percentage score indicates the quality of the article clusters to accommodate at most one target-task, which is computed as the ratio of total number of negative targets to the total number of targets (-1 and +1) as given in Eq~\ref{eq8}. A higher value indicates the results provided by the concerned cluster are concise and most relevant. 
\begin{equation}
Q_c(T)=\begin{cases}
\frac{n_t(-1)}{n_t(+1)+n_t(-1)}, & if 1 \leq n_t(+1) \textless N. \\
0, & \text{otherwise}.
\end{cases}		
\label{eq8}
\end{equation}
where N indicates the total number of target-tasks, $n_t(+1)$ and $n_t(-1)$ indicates the number of targets accepted and not accepted by OCSVM.

\begin{table*}[H]
	\centering
	\caption{OCSVM target-task mapping quality score per cluster on the reduced feature set.}
	\label{tab4}
\begin{adjustbox}{width=0.8 \textwidth,center}{	
	\begin{tabular}{|l|l|l|l|l|l|l|l|l|l|l|l|l|l|l|l|l|l|}
		\hline
		CA                       & DR     & \multicolumn{16}{l|}{Target mapping score of OCSVMs trained on reduced feature clusters}                    \\ \hline
		&        & 1    & 2    & 3    & 4    & 5    & 6    & 7    & 8    & 9    & 10   & 11   & 12   & 13   & 14   & 15   & 16 \\ \hline
		\multirow{4}{*}{\textit{k}-means} & PCA    & 0    & 0.78 & 0.78 & 0.44 & 0.44 & 0.44 & 0.56 & 0.67 & 0.22 & 0.33 & 0.56 & 0.56 & 0.56 & 0.56 & 0    & -  \\ \cline{2-18} 
		& t-SNE  & 0    & 0    & 0    & 0    & 0    & 0    & 0    & 0    & 0    & 0    & 0    & 0    & 0    & 0    & 0    & -  \\ \cline{2-18} 
		& ISOMAP & 0.44 & 0.56 & 0.44 & 0.67 & 0.44 & 0.44 & 0.44 & 0.56 & 0.44    & 0.11 & 0.56 & 0.44 & 0.67 & 0.44 & 0.44 & -  \\ \cline{2-18} 
		& UMAP   & 0    & 0.78 & 0.67 & 0.56 & 0.56 & 0.44 & 0.67 & 0.56 & 0.33 & 0.22 & 0.56 & 0    & 0.56 & 0.11 & 0    & -  \\ \hline
		\multirow{4}{*}{DBSCAN}  & PCA    & 0    & 0    & 0    & 0    & 0    & 0    & 0    & 0    & 0    & 0.89 & 0    & 0    & 0    & 0.22 & 0    & 0  \\ \cline{2-18} 
		& t-SNE  & 0    & 0    & 0    & 0    & 0    & 0    & 0    & 0    & 0    & 0    & 0    & 0    & 0    & 0    & 0    & 0  \\ \cline{2-18} 
		& ISOMAP & 0    & 0    & 0    & 0    & 0    & 0    & 0    & 0    & 0    & 0.67 & 0    & 0    & 0    & 0.69 & 0    & 0  \\ \cline{2-18} 
		& UMAP   & 0    & 0    & 0    & 0    & 0    & 0    & 0    & 0    & 0    & 0.89 & 0    & 0    & 0.22 & 0.89 & 0    & 0  \\ \hline
		\multirow{4}{*}{HAC}     & PCA    & 0.22 & 0.56 & 0.11 & 0.78 & 0.22 & 0.67 & 0.67 & 0.11 & 0.56 & 0.44 & 0.56 & 0.89 & 0.56 & 0.22 & -    & -  \\ \cline{2-18} 
		& t-SNE  & 0    & 0    & 0    & 0    & 0    & 0    & 0    & 0    & 0    & 0    & 0    & 0    & 0    & 0    & -    & -  \\ \cline{2-18} 
		& ISOMAP & 0.44 & 0.56 & 0.11 & 0.44 & 0.44 & 0.44 & 0.56 & 0.44 & 0.44 & 0.44 & 0.44 & 0.44 & 0.44 & 0    & -    & -  \\ \cline{2-18} 
		& UMAP   & 0.56 & 0.67 & 0.22 & 0.89 & 0.22 & 0.78 & 0.67 & 0.22 & 0.67 & 0.56 & 0.56 & 0    & 0.56 & 0.22 & -    & -  \\ \hline
	\end{tabular}}
	\end{adjustbox}
\end{table*}

\begin{table*}[]
	\centering
	\caption{Overall behaviour of all models.}
	\label{tab5}
\begin{adjustbox}{width=0.8 \textwidth,center}{
	\begin{tabular}{|p{0.44in}|p{0.4in}|l|l|l|l|l|l|l|l|l|}
		\hline
		Clustering scheme   & DR scheme  & TP    & TN    & FP   & FN   & Precision & Accuracy & Recall & Specificity & F1-Score \\ \hline
		\multirow{4}{*}{Kmeans} & PCA                             & 19464 & 25748 & 4097 & 691  & 0.826     & 0.904    & 0.966  & 0.863       & 0.890    \\ \cline{2-11} 
		& t-SNE                           & 19640 & 26028 & 4101 & 411  & 0.827     & 0.910    & 0.980  & 0.864       & 0.897    \\ \cline{2-11} 
		& ISOMAP                          & 18485 & 25699 & 5076 & 740  & 0.785     & 0.884    & 0.962  & 0.835       & 0.864    \\ \cline{2-11} 
		& UMAP                            & 18907 & 26118 & 4654 & 321  & 0.802     & 0.901    & 0.983  & 0.849       & 0.884    \\ \hline
		\multirow{4}{*}{DBSCAN} & PCA                             & 23519 & 25862 & 42   & 577  & 0.998     & 0.988    & 0.976  & 0.998       & 0.987    \\ \cline{2-11} 
		& t-SNE                           & 19460 & 26028 & 0    & 411  & 1.000     & 0.991    & 0.979  & 1.000       & 0.990    \\ \cline{2-11} 
		& ISOMAP                          & 23538 & 26071 & 23   & 368  & 0.999     & 0.992    & 0.985  & 0.999       & 0.992    \\ \cline{2-11} 
		& UMAP                            & 23499 & 26128 & 62   & 311  & 0.997     & 0.993    & 0.987  & 0.998       & 0.992    \\ \hline
		\multirow{4}{*}{HAC}    & PCA                             & 19334 & 25315 & 4227 & 1124 & 0.821     & 0.893    & 0.945  & 0.857       & 0.878    \\ \cline{2-11} 
		& t-SNE                           & 18722 & 26016 & 4839 & 423  & 0.795     & 0.895    & 0.978  & 0.843       & 0.877    \\ \cline{2-11} 
		& ISOMAP                          & 19356 & 25723 & 4205 & 716  & 0.822     & 0.902    & 0.964  & 0.859       & 0.887    \\ \cline{2-11} 
		& UMAP                            & 17653 & 26121 & 5908 & 318  & 0.749     & 0.875    & 0.982  & 0.816       & 0.850    \\ \hline
	\end{tabular}}
	\end{adjustbox}
\end{table*}
Based on the extensive trials, it is observed that \textit{k}-means and HAC outperf- ormed the DBSCAN with a significant margin in generating good quality of clusters, whereas \textit{k}-means performed better than HAC as observed from the statistics in Table~\ref{tab3}, which is utilized for training the OCSVMs with the generated clusters for mapping the target domain appropriately.

The confusion matrix shown in Fig.~\ref{fig7} represents the average performance of the trained OCSVMs by using the clusters generated from \textit{k}-means, DBSCAN, and HAC in original feature space. It shows that the false negative rate is almost negligible, and ensures the robustness of the approach. Results show that among all three clustering techniques DBSCAN sounds most promising, however due to its heavily skewed distribution of articles in the clusters it is not a robust technique for this problem, hence \textit{k}-means is observed most suitable clustering technique to work along with parallel OCSVMs.

Furthermore, the complete procedure was repeated with the dimensionally reduced embedding of the articles with the concern to achieve similar or better results with less computational cost. Table~\ref{tab4} describes the target mapping results on the articles vector space whose embedded feature set is reduced from 150 to 2 dimensions. The performance of the proposed approach is analysed in the presence of reduced features as shown in Table~\ref{tab5}. The results exhibit that in presence of reduced features obtained by ISOMAP, the \textit{k}-means along with OCSVMs method outperforms others and needs less computation power.

\section{Conclusion}
The pandemic environment of COVID-19 has brought attention to the research community. Many research articles have been published concerning the novel coronavirus. The CORD initiative has developed a repository to record the published COVID-19 articles with the update frequency of a day. This article proposed a novel bottom-up target guided approach for mining of COVID-19 articles using parallel one-class support vector machines which are trained on the clusters of related articles generated with the help of clustering approaches such as \textit{k}-means, DBSCAN, and HAC from the embedded vector space of COVID-19 articles generated by doc2vec approach. With extensive trials, it was observed that the proposed approach with reduced features produced significant results by providing quality of relevant articles for each discussed target-task, which is verified by the cosine similarity score. The domain of this article is not limited to mining the scholarly articles, where it can further be extended to other applications concerned with the data mining problems.


\section*{Acknowledgements}
We are very grateful to our institute, Indian Institute of Information Technology Allahabad (IIITA), India and Big Data Analytics (BDA) lab for providing the necessary infrastructure and resources. We would like to thank our supervisors and colleagues for their valuable guidance and suggestions. We are also very thankful for providers of CORD-19 dataset.


\bibliographystyle{cas-model2-names}
\bibliography{cas-refs}

\end{document}